\documentclass[letterpaper, 10 pt, conference]{ieeeconf}  

\IEEEoverridecommandlockouts                              

\usepackage{graphics} 
\usepackage{epsfig} 
\usepackage{mathptmx} 
\usepackage{times} 
\usepackage{amsmath} 
\usepackage{amssymb}  
\usepackage[dvipsnames]{xcolor}
\usepackage[noadjust]{cite}
\usepackage{tabularx}

\newcommand{\figref}[1]{Fig.~\ref{#1}}     
\newcommand{\tableref}[1]{Table~\ref{#1}}

\newcommand{\tm}[1]{{\textsuperscript{\tiny TM} #1}}
\newcommand{\reg}[1]{{\textsuperscript{\scriptsize \textregistered} #1}}

\title{\LARGE \bf
Deep Learning Classification of Touch Gestures \\ Using Distributed Normal and Shear Force
}

\author{Hojung Choi$^{1}$, Dane Brouwer$^{1}$, Michael A. Lin$^{1}$, Kyle T. Yoshida$^{1}$, Carine Rognon$^{2}$, \\ Benjamin Stephens-Fripp$^{2}$, Allison M. Okamura$^{1}$, Mark R. Cutkosky$^{1}$
\thanks{$^{1}$Stanford University, USA
        {\tt\footnotesize \{hjchoi92, daneb, mlinyang, kyle3, aokamura, cutkosky\}@stanford.edu}}%
\thanks{$^{2}$Reality Labs Research, Meta Platforms Inc., Redmond, USA
        {\tt\footnotesize \{carinerognon, bensfripp\}@fb.com}}%
}

\begin{document}

\maketitle
\thispagestyle{empty}
\pagestyle{empty}

\begin{abstract}
When humans socially interact with another agent (e.g., human, pet, or robot) through touch, they do so by applying varying amounts of force with different directions, locations, contact areas, and durations. While previous work on touch gesture recognition has focused on the spatio-temporal distribution of normal forces, we hypothesize that the addition of shear forces will permit more reliable classification. We present a soft, flexible skin with an array of tri-axial tactile sensors for the arm of a person or robot. We use it to collect data on 13 touch gesture classes through user studies and train a Convolutional Neural Network (CNN) to learn spatio-temporal features from the recorded data. The network achieved a recognition accuracy of $74\%$ with normal and shear data, compared to $66\%$ using only normal force data. Adding distributed shear data improved classification accuracy for 11 out of 13 touch gesture classes.

\end{abstract}

\section{Introduction}

 Touch gestures play a crucial role in communicating emotions and building relationships \cite{field2014touch}. The high-dimensional combinations of forces, contact locations, and timing characteristics are integrated to create a socially defined meaning in a given context \cite{Yohanan2011}. For instance, a stroke on the arm of a friend in distress can convey emotional support without oral communication \cite{mcintyre2021language}. As remote communication such as video chats or virtual environments \cite{Kim2022Metaverse} become more commonplace, there is a growing need to replicate this dimension of interaction in a distant setting because a lack thereof can cause loneliness \cite{rognon2021online}. Looking toward the future, it is not difficult to imagine interacting with robots through touch; an anthropomorphic home robot may receive a
 gentle pat on the back after a cleaning job, or could respond to a cautionary tap if about to make a mistake. In such cases, it is imperative to capture the physical information embedded in a touch gesture to recognize what type of gesture it is.
 
 Previous work on touch gesture recognition has incorporated various types of sensors. 
 Tawil \emph{et al.}~\cite{SilveraTawil2012} designed an electrical impedance tomography (EIT) based sensor to recognize normal force magnitude and contact location along the arm of a mannequin. They used the \emph{LogitBoost} algorithm to classify touch gestures into 9 different classes. Their more recent work expands to classifying underlying emotions from touch gestures \cite{SilveraTawil2014}, a topic that has recently received increasing interest \cite{SalvatoTOH2021, mcintyre2021language}. Jung \emph{et al.}~\cite{Jung2014} wrapped the forearm of a mannequin with discrete piezoresistive sensors to sense normal pressure in multiple contact areas and create a touch gesture dataset. Datasets were used further for advanced classification algorithms such as Neural Nets \cite{vanWingerden2014} or Random-Forest and Boost \cite{gaus2015social} with hand-designed features, increasing accuracy and analyzing principal features for classification. 

Displaying touch gestures recorded from tactile sensors to a remote touch receiver has also been studied. Huisman \emph{et al.}~\cite{Huisman2013} designed TaSST, an integrated wearable pressure sensor and actuator sleeve that was useful for conveying simple and protracted touches. Salvato \emph{et al.}~\cite{SalvatoTOH2021} conveyed emotion through discrete indenters playing recorded touch gestures. These investigations focused on measuring and actuating normal forces, omitting the shear information that may arise, for example, from a \emph{stroke} or \emph{pull}. Such gestures particularly motivate the addition of shear data.

\begin{figure}[tb!]
\centering
	\vspace{1.5mm}
	\includegraphics[width=3.4in]{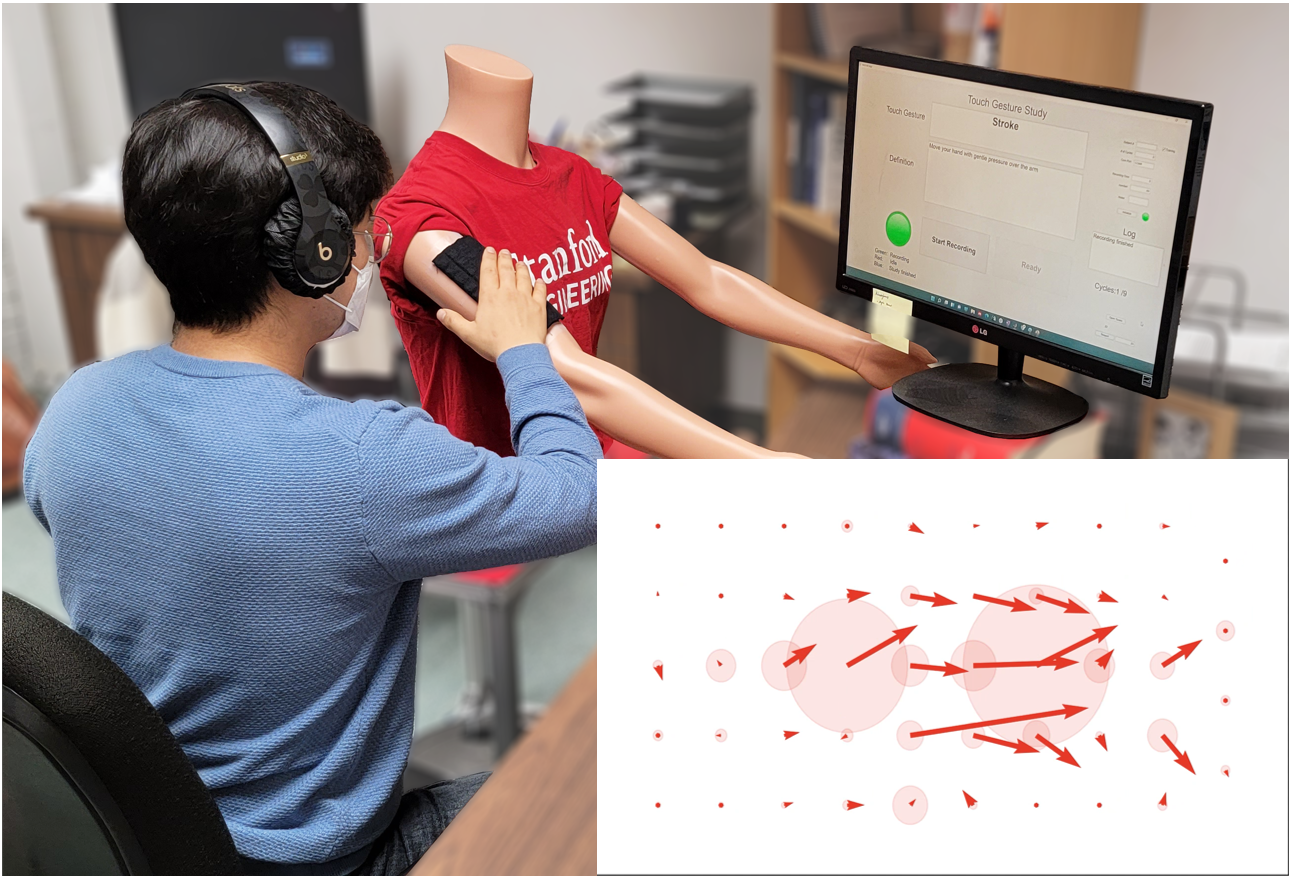}
	\caption{User study setup for collecting touch gesture data consisting of a mannequin and our flexible sensing array. The bottom right plot shows sensor data; red circle diameters and arrows indicate normal
	and shear forces, respectively.}
	\label{fig:CoverPhoto}
	\vspace{-7pt}
\end{figure}

Research on multi-axial distributed tactile arrays has produced diverse solutions, including piezoresistive \cite{Oh2020, Li2021, Wang2019}, capacitive \cite{Huh2020, Cheng2010-mr, Boutry2018}, and magnetic \cite{Tomo2016, bhirangi2021reskin, Yan2021}. Among them, magnetic approaches have recently increased in popularity due to their large signal-to-noise ratio and the availability of small tri-axial sensors suitable for inclusion in an array \cite{Tomo2016}. While arrays of magnetic sensors have been applied to robotic fingertips \cite{Tomo2018, Yan2021} or to the human arm for contact localization \cite{bhirangi2021reskin}, we believe our design, an array of tri-axial sensors in a soft sleeve-like form factor, is new.

In this paper, we introduce a magnetic, distributed, flexible, tri-axial force sensor that incorporates shear force to classify touch gestures. We designed and fabricated the tri-axial sensor and collected a touch gesture dataset by conducting a user study where participants provide gestures of 13 different types (\tableref{tab:GestureDefinitions}) to our sensor while it was wrapped on a mannequin arm (\figref{fig:CoverPhoto}). We used a Convolutional Neural Network (CNN) to learn useful spatio-temporal features from the high-dimensional data. Our results show that adding shear information increases average classification accuracy by $8\%$; using only normal force gives a $66\%$ accuracy while normal and shear combined results in a $74\%$ accuracy (\figref{fig:ConfusionMatrix}).

\section{Tri-Axial Magnetic Tactile Array}

\subsection{Sensor Design and Working Principle}

\begin{figure}[thpb]
\centering
	\vspace{1.5mm}
	\includegraphics[width=3.0in]{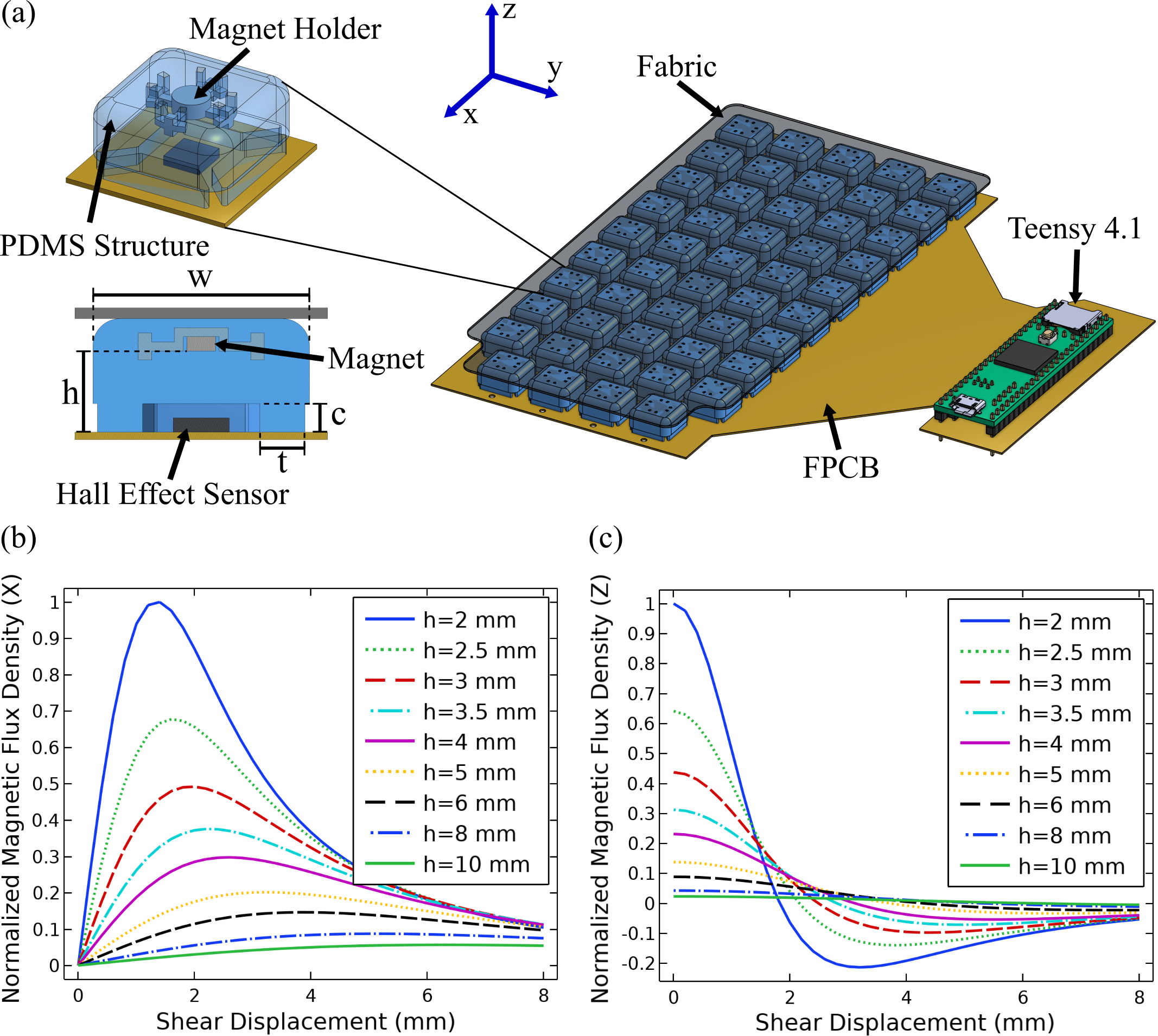}
    \vspace{+3pt}
	\caption{ Sensor design and working principle. (a) A rendering of the sensor array and an individual taxel.  (b) Horizontal and (c) vertical magnetic flux readings from FEA simulations when varying magnet heights and displacements.}
	\label{fig:SensorDesign}
	\vspace{-7pt}
\end{figure}

Our sensor is a 5 $\times$ 9 array with an additional column of 4 taxels that are each tri-axial force sensors. The footprint of the sensing area is 8\,cm $\times$ 16\,cm to cover the outer surface of the upper arm. As shown in \figref{fig:SensorDesign}(a), each taxel is a pair of a disc magnet (K\&J Magnetics, D101-N52) embedded in a PDMS structure and a hall effect sensor (Melexis, MLX90393) placed on a flexible printed circuit board (FPCB). A sheet of fabric (Needles, French Terry), bonded to the top of each taxel, provides a continuum surface for users to provide touch gestures. The taxels are located 1.5\,cm apart, which is both the average size of a human fingertip \cite{Dandekar2003} and the distance required to minimize cross-coupling. Taxels were designed to be squares to minimize gaps, creating a continuous surface.

Three axes of force are measured by detecting the change in magnetic flux densities in x, y, and z directions (\figref{fig:SensorDesign}(a)) due to the change in the relative position and orientation between the magnet and the hall effect sensor. The PDMS structure holding the magnet is a square platform with four walls underneath, resembling a hollow beam. This design maximizes translation and minimizes rotation of the magnet under shear, as rotation of the magnet can offset the signal from translation and decrease sensitivity.

TAP\reg Silicone RTV (25A-30A Shore Hardness, 600\,psi tensile strength) was selected for the PDMS structure due to its low stiffness and high tensile strength. The primary geometric parameters -- wall thickness, $t$, width, $w$, cavity height, $c$, and magnet height, $h$ -- are shown in \figref{fig:SensorDesign}(a). The wall thickness, $t$, was maximized to increase bonding area, and therefore robustness, without significantly increasing stiffness or causing collisions with the Hall Effect chip. Experimentally, a 2.5\,mm wall thickness proved effective. The structure's width, $w$, was chosen to minimize the gap between taxels without causing collisions under significant shear force. A 12\,mm width was selected, and the absence of collisions was verified empirically. The height of the cavity, $c$, was fixed at 1/16" to ease manufacturing and minimize bulk rotation under shear without colliding with the chip under significant normal force. 

To establish a suitable magnet height, the change in horizontal and vertical magnetic flux during shear displacements of a magnet at varying heights was investigated using finite element analysis (FEA) in COMSOL. The results can be seen in \figref{fig:SensorDesign}(b) and \ref{fig:SensorDesign}(c). The horizontal flux plot (\figref{fig:SensorDesign}(b)) shows that for small magnet heights, the initial shear displacements significantly change the sensed magnetic flux orientation and therefore lead to large signals. As the magnet continues to shear, the sensed magnetic field intensity decreases and is further reoriented to reduce the horizontal flux, leading to a peak in signal. This phenomenon eliminates a one-to-one mapping between magnet displacement and horizontal flux, making the interpretation of these signals ambiguous. A similar behavior is seen in the vertical flux plot (\figref{fig:SensorDesign}(c)), but the peak in signal occurs at larger shear displacements as compared to the horizontal flux. Since the gap between taxels is 3\,mm, we want to avoid this signal peak before the magnet has displaced that much. For large magnet heights, the total change in signal is significantly smaller than in small magnet heights. A magnet height, $h$, of 6\,mm was therefore selected as a balance between avoiding a signal peak and maximizing the signal. 


For system integration, the 49 Hall Effect sensors were connected to an embedded microcontroller (Teensy 4.1 Cortex M-7 @600\,MHz) on a custom FPCB. Two SPI modules connecting 24 and 25 sensors each were used to distribute the capacitive load from the sensors to the high-frequency clock line. The entire array is sampled at 25\,Hz via serial communication.

\subsection{Fabrication}

\begin{figure}[tb!]
\centering
	\vspace{1.5mm}
	\includegraphics[width=3.0in]{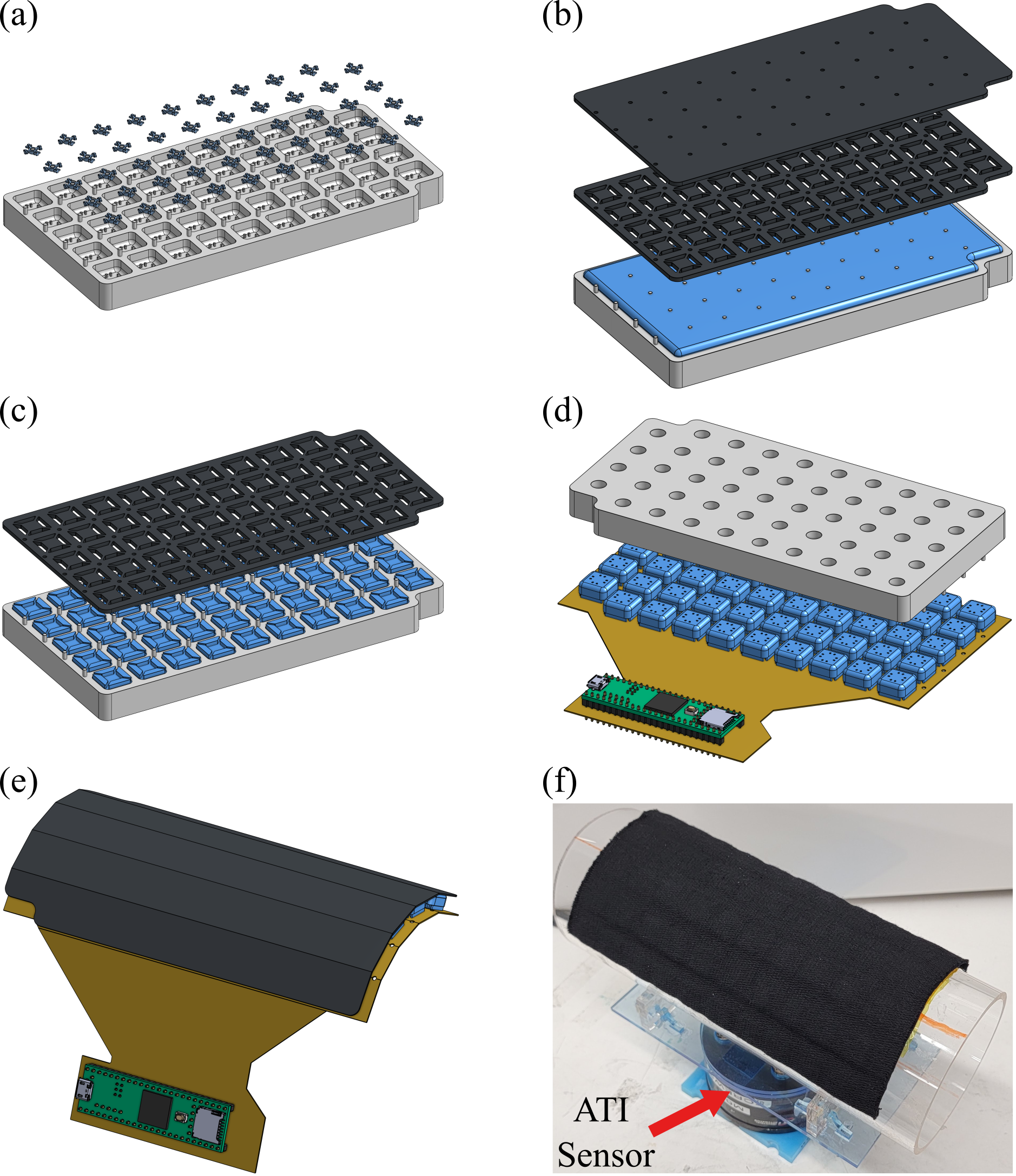}
    \vspace{+3pt}
	\caption{Sensor fabrication process. (a) Assemble magnets into magnet holders and into the mold. (b) Pour silicone into the mold and press down the mask. (c) Once cured, remove the mask and place a thin layer of fresh silicone onto the base of the walls. (d) Press silicone structures onto a primed FPCB using alignment pins and allow to cure. (e) Pre-curve sensor to desired configuration and adhere fabric by applying a thin layer of Sil-Poxy\tm on top of each taxel. (f) The curved sensor is placed on top of an ATI-Gamma force/torque sensor for calibration.}
	\label{fig:SensorFabrication}
	\vspace{-7pt}
\end{figure}

The sensor array is fabricated according to the process illustrated in \figref{fig:SensorFabrication}. First, the molds for the PDMS structure and magnet holders are printed on a StrataSys\reg Objet24\tm 3D printer on VeroWhitePlus\tm material and the masks are laser cut from 1/16" acrylic using a Universal\reg Laser Systems PLS6.150D laser cutter. Each magnet is then bonded to a magnet holder using LOCTITE\reg 401\tm Instant Adhesive. Smooth-On\reg Universal\tm Mold Release is applied to the mold and masks to ease demolding. Each of the magnet holder assemblies is then placed into the corresponding cavity of the mold on top of alignment pins and is held in place using a magnet on the opposite side of the mold. TAP\reg Silicone is mixed according to instructions with 5\% by weight of Smooth-On\reg Silicone Thinner\tm then poured into the mold and degassed. Next, the mask is placed on top of the uncured silicone, allowing the silicone to flow up and create the structure's walls. A flat piece of acrylic and weights are added on top, then it is placed into a pressure chamber at 60\,kPa overnight to cure.  

Once the silicone is cured, the mask is removed, and any excess films are trimmed from the structures. Next, a thin layer of fresh silicone is added to the base of the structure's walls by spreading silicone onto a flat piece of acrylic and dipping the mold onto it. The alignment pins on the mold are then used to place the structures onto an FPCB that has been gently abraded using 400 grit sandpaper and primed using DOWSIL\tm PR-1200 RTV Primer. A weight is placed atop the mold, and the bonding layer of silicone is allowed to cure. Upon removal of the mold, the tops of each taxel are cleaned with isopropyl alcohol and the sensor is curved around an acrylic cylinder (8\,cm in diameter) to approximate the final configuration on the mannequin arm. The top of each taxel is then coated with a thin layer of Smooth-On\reg Sil-Poxy\tm to bond the fabric to the sensor array. Pre-curving the sensor prevents excessive strain of the fabric, allowing the final array to conform to an arm with minimal change in signal. Once the Sil-Poxy\tm cures, the sensor is ready for calibration.

\subsection{Sensor Characterization}

\begin{figure}[tb!]
\centering
	\vspace{1.5mm}
	\includegraphics[width=3.4in]{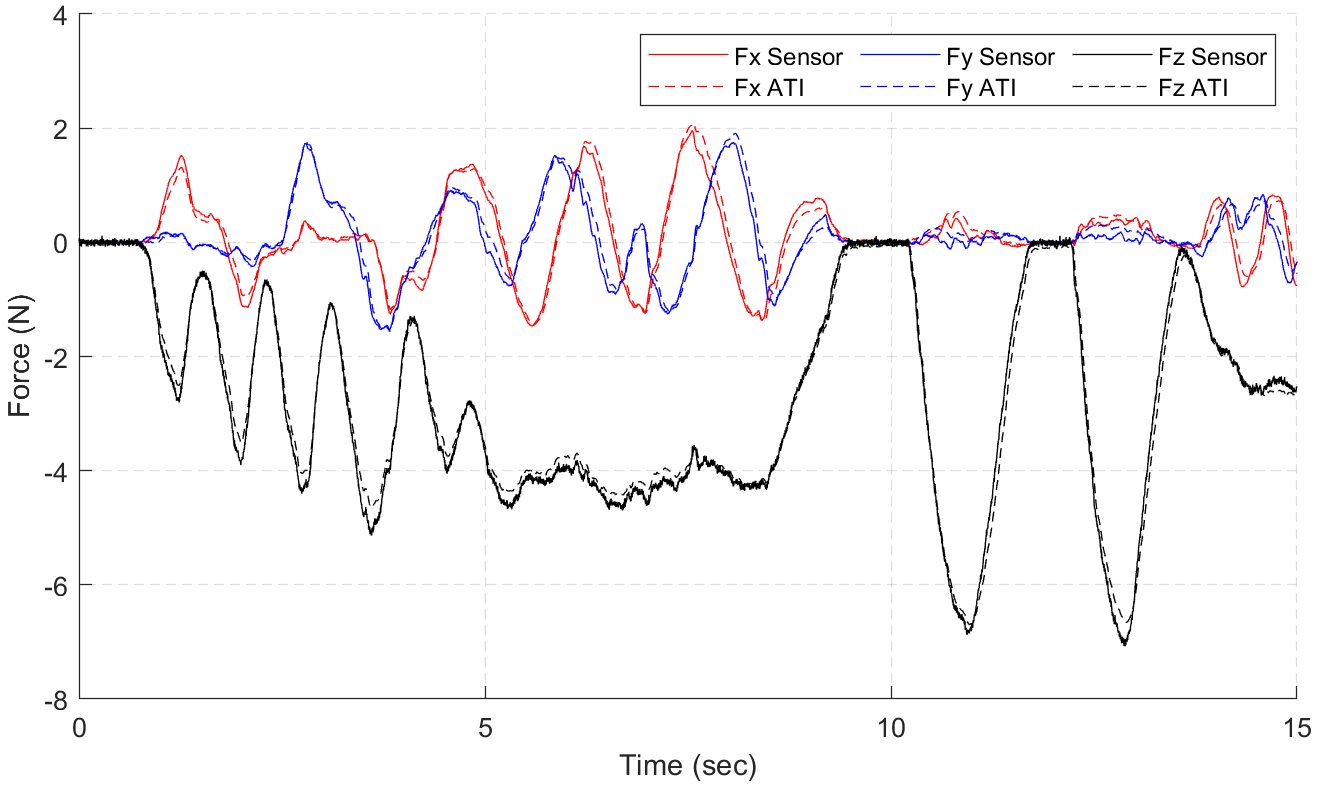}
	\caption{Calibrated sensor and reference force/torque data match closely.}
	\label{fig:SensorCalibrationGraph}
\end{figure}

\begin{figure*}[tb!]
\centering
	\vspace{1.5mm}
	\includegraphics[width=7.0in]{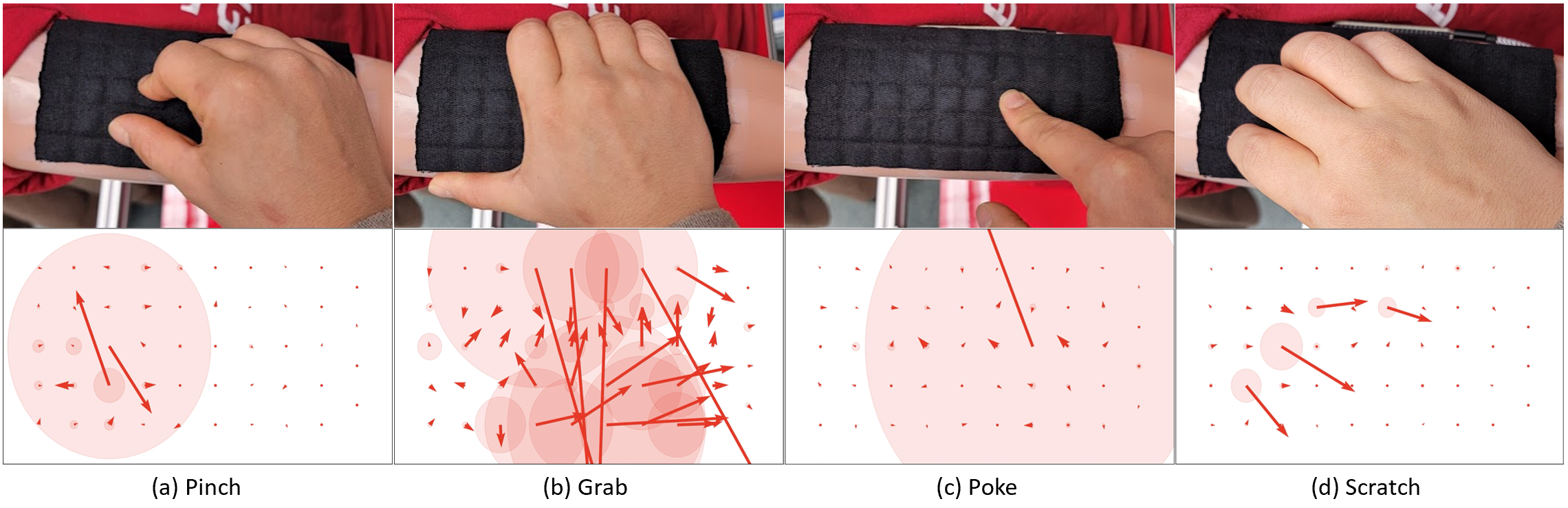}
	\caption{Visualization of the distributed normal and shear forces for selected touch gestures. The diameter of the semi-transparent red circle indicates the normal force. The magnitude and direction of the red arrows indicate the shear force.}
	\label{fig:TouchPlots}
	\vspace{-7pt}
\end{figure*}

The 49 taxels were calibrated individually using a 2nd-order least squares fit without any bias term to ensure 0 reading under no force input. The acrylic cylinder with the sensor attached on its surface was mounted on a reference force/torque (F/T) sensor (ATI, Gamma SI-32) which provides the ground-truth force. The top surface of the taxels to be calibrated were oriented parallel to the surface of the reference F/T sensor in order to align the coordinate axes. The force inputs were provided through the experimenter's fingertip and moment information was excluded from calibration (Input ranges $F_{x, y}$: $\pm$2\,N, $F_z$: 0 $\sim$ -7\,N). \tableref{table:ErrorTable} shows the overall performance of all 49 taxels. 
\figref{fig:SensorCalibrationGraph} is an example of the fit to unseen data, demonstrating that the force reading from each taxel tracks the ground-truth information. The visualization of the entire array under common touch gestures is presented in \figref{fig:TouchPlots}, which shows that the proposed sensor captures salient features such as touch location, normal force, and shear force well with negligible cross-coupling between taxels.

\begin{table}[tb!]
    \small
	\centering
	\renewcommand{\arraystretch}{1.2}
	\caption{Mean and Standard Deviation of RMS error for calibration. (Input ranges $F_{x, y}$: $\pm$2\,N, $F_z$: 0 $\sim$ -7\,N)}
	
	\begin{center}
	\begin{tabularx}{0.4\textwidth}{c c c c}
    

	Input data set     & $F_x$ $(N)$  & $F_y$ $(N)$  & $F_z$ $(N)$  \\
	\hline
	Mean               & 0.1411 & 0.1336 & 0.1896 \\
	Standard Deviation & 0.0290 & 0.0262 & 0.0534 \\

	\hline
	\label{table:ErrorTable}

	\end{tabularx}
    \vspace{-20pt}
\end{center}

\end{table}

\section{Deep Learning Classification \\ of Touch Gestures}

During daily interactions, humans differentiate touch gestures based on features such as the location and area of contact, applied normal and shear forces, and duration. Inspired by the growing popularity of applying deep learning in tasks that involve classification of data with high-dimensional features \cite{Krizhevsky2012AlexNet}, we employ Convolutional Neural Networks (CNN) to train a generalizable model for touch gesture classification. We hypothesize that this is a more effective method than previously reported approaches where features were designed manually \cite{vanWingerden2014} because CNNs are capable of learning useful features for classification from data. In particular, we show the importance of incorporating shear force information by performing an ablative study with and without shear force. 

\subsection{User Study}

A user study was conducted to collect touch gesture data. Eleven right-handed participants (5 male, 6 female) were recruited through Stanford University email lists that include students in various departments. Participants provided informed consent, and the protocol was approved by the Stanford University Institutional Review Board (Protocol \#26526). A subset of the touch dictionary described in \cite{Yohanan2011} consisting of 13 touch gestures was chosen as our touch class, as shown in  \tableref{tab:GestureDefinitions}. The subset was chosen based on whether a gesture can be applied to the upper arm and does not have explicitly overlapping definitions (such as \emph{squeeze} instructing users to ``press"). As shown in \figref{fig:CoverPhoto}, the setup consists of our sensor mounted on the right upper arm of a mannequin fixed on an extruded aluminum frame. The user is guided to apply different touch gestures using their right hand through the Graphical User Interface (GUI).

\begin{table}[h]
    \caption{Touch Gesture Classes}
    \label{tab:GestureDefinitions}
    \renewcommand{\arraystretch}{1.2}
    \begin{center}
        \begin{tabular}{ | c || m{6cm} | }
        \hline
        \textbf{Class} & \textbf{Touch Gesture Definition} \\
        \hline
        \hline
        Stroke & Move your hand with gentle pressure over the arm \\
        \hline
        Scratch & Rub the arm with your fingernails \\
        \hline
        Tickle & Touch the arm with light finger movements \\
        \hline
        Pat & Gently and quickly touch the arm with the flat of your hand \\
        \hline
        Tap & Strike the arm with a quick light blow or blows using one or more fingers \\
        \hline
        Slap & Quickly and sharply strike the arm with your open hand \\
        \hline
        Poke & Jab or prod the arm with your finger \\
        \hline
        Pinch & Tightly and sharply grip the arm between your fingers and thumb \\
        \hline
        Pull & Exert force on the arm by taking hold of it in order to move it towards yourself \\
        \hline
        Rub & Move your hand repeatedly back and forth on the arm with firm pressure \\
        \hline
        Press & Exert a steady force on the arm with your flattened fingers or hand \\
        \hline
        Grab & Grasp or seize the arm suddenly and roughly \\
        \hline
        Shake & Move the arm up and down or side to side with rapid,  forceful, jerky movements \\
        \hline
        \end{tabular}
        \vspace{-7pt}
    \end{center}
\end{table}

To create a dataset representative of touch gestures from various individuals, the user study protocol is designed such that it ensures consistency in the way participants understand what each touch gesture means while preserving user-to-user variability in the way gestures are applied.
The study starts with a familiarization stage designed to mitigate the effects of misunderstanding users may have about each touch gesture. For instance, \emph{tap} and \emph{pat} may be used interchangeably in our daily lives, while there is a clear difference in the definition for our study; \emph{tap} limits contact only to the fingers while \emph{pat} extends to the palm, as shown in \tableref{tab:GestureDefinitions}. Users are shown text definitions and two video demonstrations corresponding to each touch gesture. Videos, included in the supplementary material, were also provided as pilot studies revealed that participants tend to fixate on following the text by the word rather than applying their usual gesture. To minimize bias, the touch gesture in each video example was performed on the shoulder or the back, while users were prompted to contact the upper arm. Additionally, the frequency and duration of the touch gesture in each video was different, allowing users to perform touch gestures most naturally rather than be constrained by any requirements. The first video shows one person applying a touch gesture to another person, where the first person is prompted by the touch gesture definitions in \tableref{tab:GestureDefinitions}. The second video shows the hand of one person providing a touch gesture to another person using his non-dominant, left hand. Users were then given a chance to apply the touch gesture on the sensor surface before familiarizing themselves with the next gesture.

During the data collection stage, users were prompted to provide each touch gesture three times in a pseudo-randomized order which completes a single block. Each block is pseudo-randomized differently, and one user study consists of 9 blocks. The repetition of gestures to form blocks accelerates the data collection process and limits user fatigue. Users were given a break up to 2 minutes every 3 blocks. Each touch was recorded by the sensor for 5 seconds, in which users were given the freedom to choose the number or duration of the prompted touch gesture. We collected 3861 time series of touch gesture data from 11 users.

\subsection{Convolutional Neural Network Model}

\begin{figure}[b!]
\centering
	\includegraphics[width=3.4in]{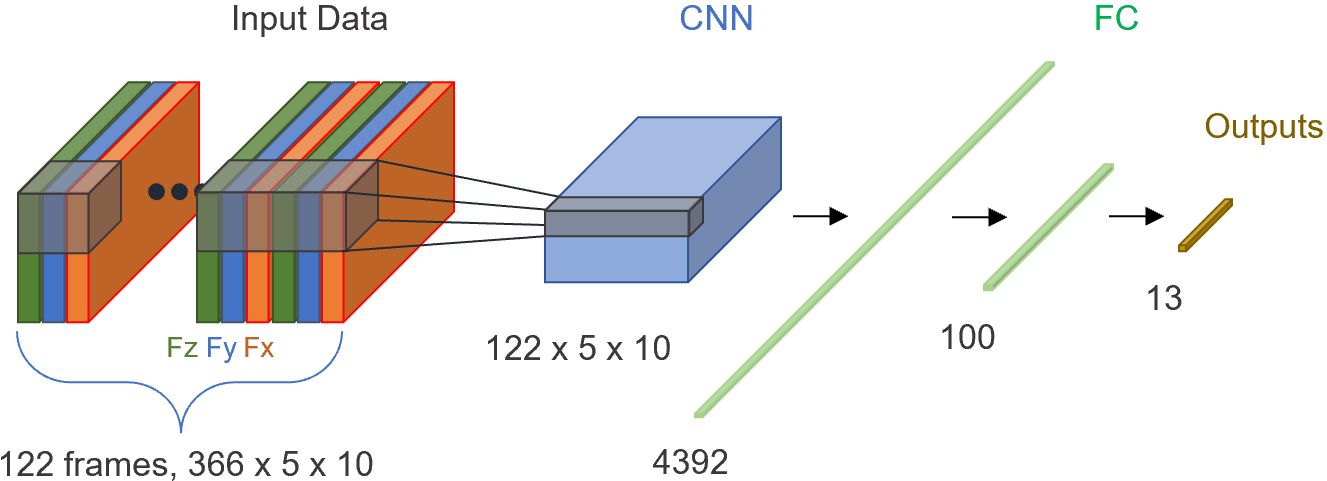}
	\caption{CNN model used for touch gesture classification. Tri-axial force data are stacked along the temporal length to be treated as a 3D image. FC stands for a fully connected layer.}
	\label{fig:CNN}
\end{figure}

The touch gesture data collected from our sensor is analogous to a 5 second video where each sensor taxel is an image pixel. High level features of the gesture such as area of contact, number of contacts, and magnitude and direction of force along with their temporal development is embedded. In order to extract useful information, we designed a CNN classifier that can consider spatio-temporal features by treating the data as a 3D image (\figref{fig:CNN}). 

The tri-axial force information from 49 taxels is converted to a $3 \times 5 \times 10$ $[Channel \times Height \times Width]$ image, by padding zeros to the imaginary taxel locations where there are only 4 taxels in the column. Each of the frames from the force video is then stacked along the channel axis which results in a $366 \times 5 \times 10$ tensor as each recording is 122 frames long. The entire dataset of 3861 recordings was then split into $3081 / 390 / 390$ for training, validation, and test sets respectively, evenly across users. With $3081$ touch gesture recordings for training, the input to the CNN was a $3081 \times 366 \times 5 \times 10$ tensor when all tri-axial force data were considered, and $3081 \times 122 \times 5 \times 10$ for the ablative study considering only normal force. To facilitate training, the tri-axial forces were normalized. Zero padding of size 1 has been applied to better utilize the features on the edge of the videos.

As shown in \figref{fig:CNN}, the classifier consists of 1 convolution layer with a kernel size of 3 and stride of 1 followed by a ReLU activation, then a dropout layer (p = 0.5), and a max pooling layer with kernel size of 2 and stride of 1. This resulted in 4392 learned features which were then passed to a fully connected layer with a ReLU activation, which reduced the feature size to 100. These 100 features were fed through another fully connected layer to output the 13 predictions corresponding to the 13 different touch gesture classes. An Adam optimizer with a learning rate of 0.0001 and cross entropy loss function was employed.

\section{Results and Discussion}

\begin{figure*}[thpb]
\centering
	\vspace{1.5mm}
	\includegraphics[width=7.0in]{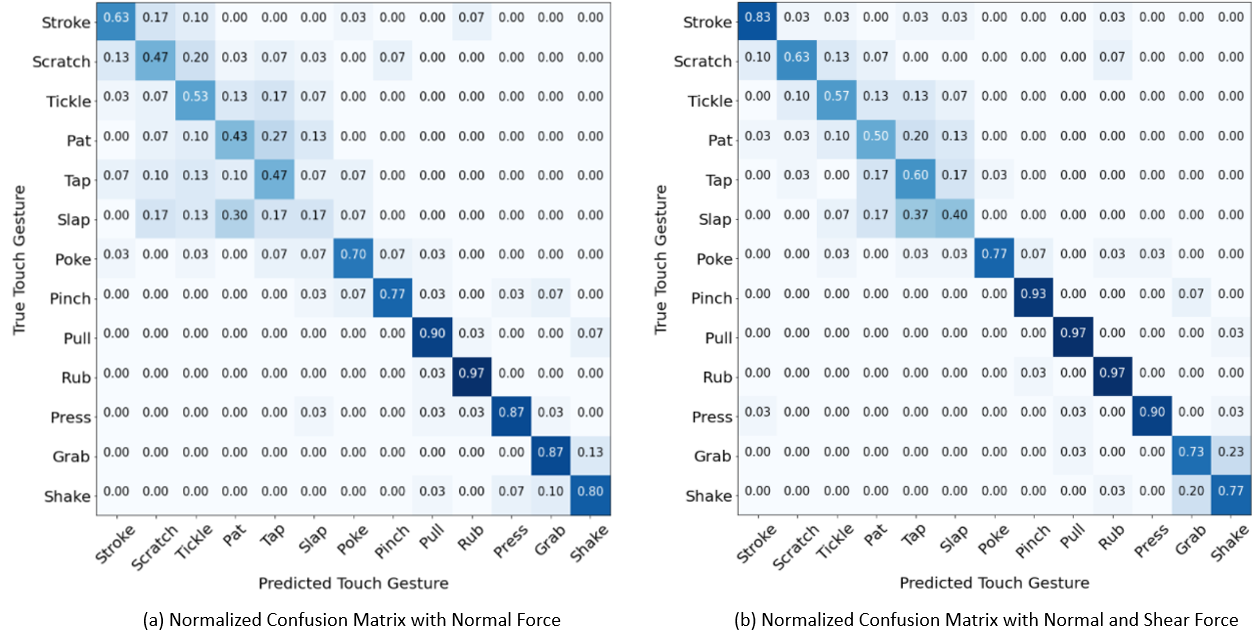}
	\caption{Results of the CNN ablative study (a) when normal force is used (average accuracy: $66\%$) and (b) when normal and shear force are used (average accuracy: $74\%$) for classification. }
	\label{fig:ConfusionMatrix}
	\vspace{-7pt}
\end{figure*}


The classification results on the ablative study are shown in \figref{fig:ConfusionMatrix}. The average accuracy for considering both normal and shear force is $74\%$, while using only normal force gives $66\%$, showing that considering shear force improves the classification of touch gestures.

Shear force greatly helped decrease confusion among \emph{stroke}, \emph{scratch}, and \emph{tickle}. This may be due to the overlapping normal force characteristics of the three gestures where in many cases there was a repetitive unidirectional movement of a round or line contact patch (Figs. \ref{fig:CoverPhoto} and \ref{fig:TouchPlots}). With shear force included, features such as the ratio between normal and shear force of these gestures may help with classification. For instance, scratch may have a higher shear to normal ratio.

Accuracy is higher for all touch gestures when shear force is considered except for \emph{grab} and \emph{shake}. The addition of shear force helps differentiate touch gestures that have similar normal force profiles. For instance, \emph{poke} and \emph{pinch} is easier to classify when the two opposing shear forces for \emph{pinch} are considered as shown in \figref{fig:TouchPlots}(a) and \ref{fig:TouchPlots}(c). The decrease in accuracy for \emph{grab} and \emph{shake} with shear force may be due to a overall similar force profile of the two gestures. Grab shows an increase in normal force in a large area with opposing shear distribution as illustrated in \figref{fig:TouchPlots}(b) and oscillation is added to a similar profile for shake. The confusion may come from how the mannequin oscillates after initial impact after \emph{grab} and that the CNN is less capable of understanding the temporal development of features as compared to sequential models such as Long Short-Term Memory (LSTM) \cite{Hochreiter1997LSTM} or Transformers \cite{Vaswani2017Transformer}.

Adding shear information did not notably improve accuracy for gestures such as \emph{press}, \emph{rub}, \emph{pull}, and \emph{poke}; their accuracy using just normal force was already high ($>70\%$). Both \emph{press} and \emph{poke} can be characterized principally by normal force, while there can be shear components depending on the angle at which the gesture was given. While rich shear components are present in \emph{rub}, it being the only periodic movement along the sensor among the gestures may make it unique and easy to classify. Just observing the change in directions of a normal force patch moving may be enough to distinguish it from similar motions such as \emph{stroke}. This could be evidence that our network is capable of considering temporal development of the force data. \emph{Pull} was hypothesized to show the largest improvement in accuracy with the addition of shear force as it may be confused with \emph{press} otherwise, but this was not the case. This may be due to the cylindrical surface of the mannequin arm since users mostly applied force near the center of the sensor for \emph{press} but grabbed the inner side of the cylindrical surface and applied a combination of normal and shear force for \emph{pull}, resulting in a pressure gradient. Thus, \emph{press} and \emph{pull} may have been distinguishable just using normal force.

For both cases of the ablative study, a significant confusion was observed across classes \emph{tickle}, \emph{pat}, \emph{tap}, and \emph{slap}. Among them, it was especially difficult to classify among \emph{pat}, \emph{tap}, and \emph{slap}. This was expected since during the user study the magnitude at which users provided these gestures was highly variable. For example, some users performed the \emph{pat} gesture using larger force than the \emph{slap} of other users. Despite the familiarization process, users seemed to rely on their preconception of the gestures as the study progressed, varying from the text and visual instructions and causing difficulty for the CNN model to classify. \emph{Tickle} was also confused with \emph{pat}, \emph{tap}, and \emph{slap} because some users tapped or patted multiple times, leading to similarities to the force profiles of \emph{tickle}. Even with such confusion, the CNN model performed better overall with the inclusion of shear force. As users do not always apply these gestures strictly along the normal axis, the extra information coming from shear force helps with classification. For instance, the conspicuous increase in accuracy for \emph{slap} may be due to the high shear force content in many slaps.




While the overall accuracy of $74\%$ when shear force is included is much higher than chance ($7.7\%$), it is lower than other applications of CNN on images \cite{Krizhevsky2012AlexNet}. This is due to the user-to-user variability in providing touch gestures and a limited dataset size. Users demonstrated a high variation on the location, number, duration, and magnitude of different forces for the same touch gesture class. Some users even misunderstood the touch definitions completely and provided a gesture for another class which resulted in mislabelled data. These errors were also considered to be user-to-user variability and were not corrected. These factors are reasons behind the comparatively low classification accuracy of touch gestures as also noted in \cite{SilveraTawil2012,gaus2015social}. 



\section{Conclusions and Future Work}

In this paper, we introduce the use of shear force for touch gesture classification. A custom-designed, magnetic, tri-axial, flexible force sensor array with high accuracy was developed. It was placed on a mannequin arm as a data collection setup. Touch gesture data were taken from 11 users yielding 3861 sets of time series data. The data were used to develop a CNN that resulted in a $74\%$ classification accuracy on 13 touch gestures when provided both normal and shear force, an improvement from $66\%$ accuracy when only normal force was considered. This result supports our hypothesis that including shear force information helps classify touch gestures.

Although the addition of shear force allows us to perform better than other studies on touch gesture classification ~\cite{Jung2014, vanWingerden2014, SilveraTawil2012}, we are interested in further improvements. In future work, larger tactile arrays can be designed to cover the entire upper arm to capture touch gestures more comprehensively. In addition, advanced models that better consider sequential information such as LSTM and Transformers can be used to increase accuracy. Various data augmentation techniques can be applied to our dataset to improve the performance of the classification models. There may also be interesting discoveries when the data segmentation for model training is grouped based on users, not randomly and equally from all users. The tri-axial sensor array can also be integrated with an array of tri-axial actuators to accurately read touch gestures and instruct what inputs to provide to an end user. Using the integrated sensor and actuator, a user study can be expanded to present a comprehensive contextual information source to motivate natural gestures and create realistic social touch interactions.



\section*{ACKNOWLEDGMENTS}

This work was supported in part by Reality Labs Research, Toyota Research Institute, a Kwanjeong Fellowship, a Stanford Graduate Fellowship, a National Science Foundation Graduate Research Fellowship Program awarded to Stanford University, and the Stanford Center at the Incheon Global Campus. The authors thank Rachel Thomasson, Negin Heravi, Elyse Chase, Mike Salvato, and the Collaborative Haptics and Robotics in Medicine Lab at Stanford University for the exciting and helpful discussions.


\bibliographystyle{IEEEtran}
\bibliography{References}

\end{document}